\documentclass[conference]{IEEEtran}
\IEEEoverridecommandlockouts
\usepackage{cite}
\def\BibTeX{{\rm B\kern-.05em{\sc i\kern-.025em b}\kern-.08em
    T\kern-.1667em\lower.7ex\hbox{E}\kern-.125emX}}

\usepackage{algorithmic}
\usepackage{graphicx}
\usepackage{textcomp}
\usepackage{xcolor}

\usepackage{algorithm}
\usepackage{subfigure}

\usepackage{diagbox}
\usepackage{threeparttable}
\usepackage{multirow}
\usepackage{color}
\usepackage{amsmath}
\usepackage{url}

\usepackage{setspace}
\usepackage{bbold}

\usepackage{booktabs}

\usepackage{etoolbox}
\makeatletter
\patchcmd{\@makecaption}
{\scshape}
{}
{}
{}
\makeatletter
\patchcmd{\@makecaption}
{\\}
{.\ }
{}
{}
\makeatother

\begin{document}

\title{Fair Decision-making Under Uncertainty}

\author{Wenbin Zhang and Jeremy C. Weiss\\
Carnegie Mellon University, United States\\
Email:\{wenbinzhang, jeremyweiss\}cmu.edu
}

\maketitle

\begin{abstract}

There has been concern within the artificial intelligence (AI) community and the broader society regarding the potential lack of fairness of AI-based decision-making systems. Surprisingly, there is little work quantifying and guaranteeing fairness in the presence of uncertainty which is prevalent in many socially sensitive applications, ranging from marketing analytics to actuarial analysis and recidivism prediction instruments. To this end, we study a longitudinal censored learning problem subject to fairness constraints, where we require that algorithmic decisions made do not affect certain individuals or social groups negatively in the presence of uncertainty on class label due to censorship. We argue that this formulation has a broader applicability to practical scenarios concerning fairness. We show how the newly devised fairness notions involving censored information and the general framework for fair predictions in the presence of censorship allow us to measure and mitigate discrimination under uncertainty that bridges the gap with real-world applications. Empirical evaluations on real-world discriminated datasets with censorship demonstrate the practicality of our approach.

\end{abstract}


\section{Introduction}

The role of artificial intelligence is expanding and is transforming many walks of life, such as the screening of job applications, the setting of insurance rates, the targeting of advertising, the allocation of health resource and the approval of mortgage loans~\cite{beutel2019putting,meyer2018amazon,Skirpan2017}. This trend is likely to continue, and the performance of the AI systems may match or even surpass human level~\cite{grace2018will}. However, as AI permeates our lives in domains of high societal impact, there is increased concern regarding the fairness and accountability of AI-based decision-making systems, which have been voiced within and beyond the AI community~\cite{barocas2017fairness,binns2018fairness,chang2021ethical}. In a recent and widely popular investigation conducted at ProPublica, authors Chen and Wong analyzed state-of-the-art clinical prediction models, concluding that these models are biased against black patients by systematically underperforming on them even when the treatment is aimed at a particular type of cancer that disproportionately impacts them~\cite{chen2019black}.  Because AI-based decision-making can be as biased as human and can even exacerbate disparity, there is therefore an urgent need to consider fairness in AI algorithms in order to maximize AI benefits for social good.

This has led to an active area of research into quantifying and mitigating AI unfairness for the sake of providing fairness-aware decision-making systems, \emph{i.e.}, systems that are not unduly biased for or against certain individuals or social groups. Most existing fairness notions seek to evaluate the bias of a decision-making process by means of two particular forms, \emph{disparate impact} and \emph{disparate treatment}, in legal texts~\cite{barocas2016big}. While the former~\cite{agarwal2018reductions} asks for approximate parity when allocating a more favorable outcome across different demographic groups defined by sensitive attributes (\emph{e.g.}, race or gender), the latter~\cite{dwork2012fairness} guarantees similar individuals are treated similarly irrespective of sensitive characteristics. Moreover, most of the work in the existing literature tackles the fairness problem by assuming the presence of class label, in which the fairness notions are defined based on the class label either actual or predicted, and the same predictive model is trained contingent upon for new instance prediction~\cite{zhang2021fair,zhang2022longitudinal}. There is limited work focuses on \emph{censoring settings} in which the class label could be absent due to censoring on the time to an event of interest, preventing existing fairness notions and approaches from being applied.  

However, the censoring phenomenon widely exists in many real-world applications. In addition to the aforementioned discriminatory clinical prediction study in which the patient's true time to relapse or hospital discharge could be unknown for various reasons, censoring is common when profiling customers for business planning in marketing analytics, predicting criminal recidivism for recidivism prediction instruments, to name a few. Thus, it is necessary to design decision-making systems free from discrimination while considering the characteristics of censoring settings, which is underexplored and brings unique challenges: \textbf{i) The lack of fairness definitions under uncertainty.} Although more than twenty measures of fairness have been proposed in the literature, formalizing fairness is a hard topic per se and most of them necessitate the certainty on the class label which is infeasible in the censoring settings~\cite{clark2003survival}. There is no existing fairness definition explicitly involving censorship information to quantify fairness under uncertainty. \textbf{ii) Addressing fairness in the presence of censoring.} Different from the supervised fairness methods assuming the availability of class label either actual or predicted by-design, extra care must be taken to ensure the automated decision-making system is independent of sensitive attributes-based harmful stereotypes in the presence of censoring. Addressing both unfairness and censored distribution simultaneously is challenging as the censorship accompanies and complicates the biased decision regions. In addition, directly combining/transferring from respective domains is not straightforward, sophisticated design is therefore warranted. \textbf{iii) Theoretically navigating the interplay between fairness and censoring.} Existing fairness-aware studies have largely been focused on addressing and analyzing fairness in the presence of class label. However, when there is an uncertainty on the class label, existing fairness theoretical analyses cannot apply. In addition, such a theoretical analysis is also desired so as to allow for practitioners and policy makers to navigate and customize according to their business objectives~\cite{barocas2017fairness}.       
%\textbf{iii) The censored data can be skewed with various semantic meanings.} The censored data can be rarely normally distributed with many early events and relatively few late ones. In addition, it comprises typically features with distinct semantic meanings which requires explicit attention and brings additionally challenges when addressing fairness under censorship.    

To the best of our knowledge, this is the first work that can address the above challenges and provides a generic framework that specifically incorporates censoring information for fair decision-making. More specially, the novelty of this research comes from four aspects: 

\begin{itemize}
	\item We formulate a new problem of fair decision-making under uncertainty. Then, we devise corresponding fairness notions to measure unequal treatments in the presence of censorship and attribute with different semantic meanings, thus providing necessary complements to existing fairness notions in the literature. 
	\item A respective fairness-aware learner is developed for learning with censored data that are common in many real-world applications. The proposed learner specifically accounts for censored information in the model building so as to ensure accurate predictions while minimizing unfairness in censoring settings.
	\item A theoretical analysis on the interplay between fairness and censorship is conducted, which broadens the existing fairness theory to new types. Such an analysis enables model-agnostic evaluation of fairness and censorship interplay and is also of practical purpose to help practitioners navigate through their business needs.     	
	\item Quantitative and qualitative experiments on a set of real-world discriminated datasets demonstrate that the proposed fairness notations and debiasing algorithm are indeed capable of measuring and guaranteeing fairness in the presence of uncertainty. 
\end{itemize}

The remainder of this paper is organized as follows. We will start with a review of related work on fairness-aware learning and survival analysis in Section~\ref{sec:relatedWork}. Next, we will discuss the problem definition along with notations used in this work in Section~\ref{sec:notations}. In Section~\ref{sec:methodology} we propose, to the best of our knowledge, the first fairness notions and debiasing algorithm that explicitly involve censoring information for fairness under uncertainty. Our experimental methodology and results are summarized in Section~\ref{sec:expt}. We end
with our conclusions in Section~\ref{sec:con}.

\section{Related Work}
\label{sec:relatedWork}
Relevant to our work is the work that quantifies and mitigates machine learning algorithmic unfairness and work that handles with censored data, in particular analyzing censored data with decision tree models.

\subsection{Quantifying Unfairness}

While machine learning increasingly permeate facets of life, significant concerns on the unfair and discriminatory manner of ML-based systems have been voiced and observed~\cite{beutel2017data}. The machine learning community has responded by proposing a growing body of fairness notions to measure the level of discrimination along with a number of approaches to mitigate bias in order to provide fairness-aware decision-making systems~\cite{hajian2016algorithmic}. 

%\jeremy{Fix. Disparate treatment $\neq$ individual fairness.  Former is a legal term about differences in treatment, and individual fairness is a fairness notion when comparing two individuals with similar covariate profiles but different sensitive attributes.  They just happen to both be used in the cited paper.}

The broad set of existing mathematical formulations of fairness can be typically divided into two main families, individual fairness and group fairness, based on whether they evaluate fairness at the individual level or the group level, respectively~\cite{zhang2022kis}. The former definitions aim to ensure that similarly situated individuals are treated similarly. The seminal work is the notion proposed by Dwork et al.~\cite{dwork2012fairness}, which requires similarly situated individuals to receive similar probability distributions over class labels to prevent unequal treatments. One of the prerequisites of this line of work is the demanding of a task-specific similarity metric that is suitable to measure how similar two individuals are depending on the task at hand. In practice, however, such a proper similarity metric cannot be trivially specified or it is even not possible to do so. Take the individuals with censorship as an example, a proper similarity metric is hard to be specified unless disregard the censored information which is of great importance and cannot simply be ignored. The lack of task-specific similarity metric has therefore been a major obstacle for the wider adoption of individual fairness~\cite{zhang2022fairness}.

%\jeremy{same problem as above.  you can say something like ``disparate impact is typically assessed through group fairness''}

On the other hand, group fairness notations normally first specify a set of subgroups defined by sensitive attributes (\emph{e.g.}, race or gender), then preserve fairness by asking for group level approximate parity of some statistic over class labels either predicted or actual. For example, statistical or demographic parity, the representative formulation belonging to this category, measures whether the desired outcomes are equally distributed across different demographic groups~\cite{kamiran2012data}. Compared to the dependence of predicted class label of statistical parity, equalized odds~\cite{zafar2017fairness}, another representative notation, further depends on the actual class label by considering the true classification difference between different subgroup.

Although many fairness notions exist, most of them, as previously discussed, formulate fairness depend on class label either actual or predicted, thus limiting their applicability in censoring settings which is prevalent in real-world applications. Keya et al.~\cite{keya2021equitable} directly extend the existing fairness notions to the application with censoring problems, which is the single relevant effort to the best of our knowledge. However, their definitions exclude the censoring information when measuring discrimination, which could introduce substantial bias, as censored information might be of importance and cannot simply be ignored~\cite{bradburn2003survival}. In addition, a similarity metric, as one of the vital components, is required to be specified with the Euclidean distance employed in their study, which could lead to inappropriate similarity evaluation due to censorship as well as the potential distinct semantic meanings of different attributes~\cite{bechavod2020metric}. Our proposed fairness notions alleviate such limitations by explicitly involving censoring information and is free of similarity metric specification.

\subsection{Mitigating Unfairness}

The fairness definitions above could be directly used or slightly modified as a constraint or a regularizer to enforce fairness, leading to three categories of mechanisms to guarantee fairness: i) pre-processing approaches, ii) in-processing approaches and ii) post-processing approaches, depending on the intervention occurs at the input/data layer, the algorithm design or the output/results of the model, respectively.

%The first strategy, \emph{pre-processing solutions}, relies on performing different data level operations such as transformation, perturbation and augmentation to mitigate the extent of inherited bias of the data. Among the most popular methods in this category are massaging~\cite{kamiran2009classifying} and reweighting~\cite{calders2009building}. The former directly swaps the class labels of selected instances to change data distribution for the sake of balanced representation. The swapped instances are selected using a ranker based on the potential accuracy deterioration in order to minimize accuracy loss while reducing discrimination. While the latter, instead of intrusively relabeling the instances, assigns different weights to different communities to reduce discrimination. Instances belonging to the deprived group will receive higher weighs comparing to instances from the unprotected group. However, due to proxies/correlations between features, the assumption behind such type of approaches, that classifiers trained on the fairly represented data could make fair predictions, does not necessarily hold. Pre-processing methods therefore typically are not quite effective as a standalone technique unless being used in conjunction with other methods with sophisticated design. 

The first strategy, \emph{pre-processing approaches}, deals with bias in the input/data level. The underlying assumption is that in order to learn a fair classifier, the training data should be discrimination-free. Therefore, the methods in this category, try to ``correct'' the data to ensure fairness in the representation of different communities. This comprises a popular category of methods as it is model-agnostic and therefore, can be employed by any classifier. For instance, massaging~\cite{kamiran2009classifying} changes the data distribution by re-labeling some of the instances in order to neutralize discriminatory effects. The instance to be altered are those close to the decision boundary, which is provided by a ranker. The method is applicable to binary classification problems, so the re-labeling is actually label swap. Massaging might result in reverse discrimination, \emph{i.e.}, communities that were favored are now discriminated. In~\cite{vzliobaite2011handling} an extension of massaging is proposed that also deals with the problem of reverse discrimination. Reweighting~\cite{calders2009building}, on the other hand, assigns different weights to the different communities, \emph{e.g.}, the deprived community will receive a higher score comparing to the favored community. However, the underlying assumption of such type of approaches does not necessarily hold as inherit bias could still persist due to proxies/correlations between features.

%In contrast, the second category, \emph{in-processing approaches}, consists of modifying existing algorithms, usually integrating fairness as a part of the objective function through constraints or regularization, to mitigate discrimination, and is therefore algorithm-specific. As an example, the Bayesian probabilistic modeling is leveraged to estimate fairness in sparse intersectional data~\cite{foulds2020bayesian}. In~\cite{zhao2019rank}, the Mann Whitney U statistic, a popular rank-based non-parametric independence test, is leveraged to measure the correlations between class label and sensitive attribute, then further reformulates it as a new non-convex optimization problem to mitigate the inherent bias of the data. More recently, a binary submatrix denoising framework is also proposed to account for unfairness~\cite{wan2020denoising}. The common assumption hold by these methods is the certainty on the class label. On the contrary, research on fairness under uncertainty has still been limited due to the significant technical challenges when addressing unfairness and censoring distribution simultaneously~\cite{keya2021equitable}. Our work seeks to fill this gap by jointly addressing bias reduction and censoring management.   

The second category, \emph{in-processing techniques}, directly modifies the learning algorithm to ensure that it will produce non-discriminating results. Typically, these methods are algorithm-specific. As an example, the fairness gain, reflected by the reduction in discrimination, is incorporated into the splitting criterion for fair tree induction~\cite{zhang2020feat}. This model was later extended as an ensemble approach offering additional adaptability and flexibility of fairness~\cite{zhang2020flexible}. In~\cite{zhang2020online}, the Mann Whitney U statistic, a rank-based non-parametric independence test, is leveraged to measure the correlations between class label and sensitive attribute, then further reformulates it as a new non-convex optimization problem to mitigate the inherent bias of the data. The common assumption hold by these methods is the certainty on the class label. On the contrary, research on fairness under uncertainty has still been limited due to the significant technical challenges when addressing unfairness and censoring distribution simultaneously~\cite{keya2021equitable}. Our work seeks to fill this gap by jointly addressing bias reduction and censoring management.

%The last category, \emph{post-processing solution}, consists of either adjusting the decision boundary of a model or directly changing the prediction labels.~\cite{hardt2016equality} processes with additional prediction thresholds to work against discrimination while the decision boundary for the protected group is shifted based on the theory of margins for boosting~\cite{fish2016confidence}. The latter approaches pay attention to the outcome of a classifier. In~\cite{kamiran2010discrimination}, for example, relabeling is performed on selected leaves of the decision tree to decrease discrimination while minimizing the effect on predictive accuracy. We emphasize that transferring such techniques to censoring settings is not straightforward as the decision boundaries could be censored themselves due to the censored distributions in censoring settings. 

The last category, \emph{post-processing solutions}, ``corrects'' the results of a model by modifying its decision regions to ensure a fairer representation of different communities. For example, in~\cite{hardt2016equality}, the prediction thresholds are adjusted to decrease discrimination while minimizing the effect on predictive accuracy whereas in~\cite{fish2016confidence}, the authors shift the decision boundary of the deprived group to work against discrimination. In this category, one could also place method for building human-interpretable models~\cite{zeng2017interpretable}. We emphasize that transferring such techniques to censoring settings is not straightforward as the decision boundaries could be censored themselves due to the censored distributions in censoring settings.

% the decision boundary for the protected group is shifted based on the theory of margins for boosting~\cite{fish2016confidence}

\subsection{Survival Analysis}

The critical challenge of the main outcome under assessment could be unknown for a portion of the study group, deemed censoring, hinders the use of many methods of analysis. This motivates the study of survival analysis to address the problems of partially survival information access from the study cohort~\cite{clark2003survival,liu2021research,zhang2018deterministic}. 

With the ubiquitous of censored data, survival analysis has gained its popularity in applications beyond its originated medical domain ranging from customer analytics and actuaries to predictive maintenance in mechanical operations~\cite{zhang2016using,zhang2018content,zhang2021disentangled}. Among the various methods proposed for modeling censored data, the Cox proportional hazards model (CPH)~\cite{cox1972regression} is the most commonly used in which the multiplicative relation between the risk, as expressed by the hazard function and covariates is described. Under the assumption of proportional hazards, deep neural networks have recently also been employed to better encode the nonlinearity of censored data~\cite{katzman2018deepsurv,tang2021}. In contrast, another line of effort is the tree based methods~\cite{bou2011review,tang2020using}, particularity random forests due to its superior capabilities in handling nonlinear effect of variables and is free of restrictive assumptions such as proportional hazards~\cite{ishwaran2008random,zhang2018content}. A comprehensive literature survey covering recent censored data modeling effort is provided in~\cite{wang2019machine}.

Amid the popularity of survival models, same as other ML approaches, care must be taken to ensure the fairness of these models. Our work situates in this under-explored research direction to tackle fairness in the presence of censoring. Our in-processing approach incorporates a fair splitting criterion that explicitly considers censoring information into the algorithm design to guide an accuracy-driven and fairness-oriented learning procedure. Relevantly, the Cox model is modified to ensure fair risk predictions in~\cite{keya2021equitable}. Three key differences are that our model: i) does not necessitate a distance metric to be specified for the wide adoption of our method, ii) explicitly includes survival information and survival time to mitigate bias in the censoring settings, and iii) is free of hyperparameter tuning to decrease the computational requirements in practice. 

%we employ ensemble learning as a powerful tool to further enhance fair risk predictions.  while these results are encouraging  

\section{Notations and Problem Definition}
\label{sec:notations}

In the typical fairness-aware learning settings, the discriminated data $X$ normally consists of a sequence of feature represented instances $x_1, x_2, \cdots, x_n$. Among the feature representation, a special attribute $G$ is referred as the \emph{sensitive attribute} and its attribute values distinguish the discriminated community, \emph{i.e.}, the deprived group, from the privileged community, \emph{i.e.}, the favored group. In addition, instances are also described by their corresponding class labels $y_1, y_2, \cdots, y_n$. However, class labels are inaccessible in the presence of censorship. 

The discriminated and censored data, in contrast to the typical data representation, therefore further contains the survival time $T$ and an event indicator $\delta$ in addition to the observed features $x$, typically represented in the form of ($x$, $T$, $\delta$). If the event of interest has occurred, $T$ is the actual time from the individual entered the study till the time of the event occurring, and $\delta$ becomes 1 indicating certainty on the event observation; otherwise $T$ corresponds to the elapsed time between individual entered the study and last follow-up with the individual, and the event indicator $\delta$ = 0, \emph{i.e.}, the survival time is censored~\cite{miller2011survival}.

When solely focus on the survival data without fairness constraints, the data is generally considered and modeled in terms of two quantitative terms, namely the hazard function and the survival function. The former models the instantaneous rate of event occurs at a specified time $t$ condition on surviving to $t$: 

\begin{equation}
h(t|x) = \lim\limits_{\bigtriangleup t \rightarrow 0} \frac{Pr(t<T<t+\bigtriangleup t| T\geq t, x)}{\bigtriangleup t}
\end{equation}

\noindent While the latter is the probability that the event does not occur up to time $t$, and can be determined from the hazard function and vice versa: 

\begin{equation}
S(t|x) = exp(-H(t|x))
\end{equation}

\begin{equation}
H(t|x)= \int_{0}^{t} h(t|x)dt
\end{equation}

\noindent where $H(t|x)$, known as the cumulative hazard function, is the intermediate function between the hazard function and the survival function, and can be naturally interpreted as the expected number of events of interest~\cite{latouche2013competing}. 

Compared with fairness-aware learning in supervised settings, addressing discrimination bias in censoring settings leads to uncertainty on $y_1, y_2, \cdots, y_n$ which limits the applicability of the existing fairness notions. In addition, the uncertainty on $y_1, y_2, \cdots, y_n$ could also further accompany and complicate the biased decision regions. Given the discriminated and censored data $X$, the aim of fairness-aware learning under uncertainty is then to model a fair survival function $H(\cdot)$ which makes accurate predictions based on $X$ but also does not discriminate with respect to $G$ for the discriminated and censored datasets.

\section{Methodology}
\label{sec:methodology}

This section first introduces the first of its kind fairness definitions specifically accounts for censoring, then the fair splitting criterion for the discrimination-aware and censorship information involved tree induction is introduced followed by a corresponding random forests based learning algorithm for fair risk prediction. Last, the interplay between fairness and censoring is theoretically analyzed. 

\subsection{Censored Fairness Metrics}
\label{sec:fairMetrics}

The presence of censorship in data limits the applicability of commonly used fairness definitions introduced in the existing fairness-aware studies. To fill this gap, we introduce two fairness metrics specifically account for model unfairness in the present of censorship to help with practitioners and policy makers arriving to fair decisions.

 \textbf{Concordance imparity}. Motivated by the previously discussed ProPublica investigation on clinical prediction models~\cite{chen2019black}, we first propose the \emph{concordance imparity (CI)} to measure whether the model under consideration has systematically underperformed on certain population. Formally, we define the CI, mathematically represent as $C_\triangle$, as the largest deviation of discriminative abilities across different demographic groups of the model:
%$\triangle_C$  $C_\triangle$
\begin{equation}
\label{equ:ci}
	%CI= \max_{g \in G}|CF(G=g)- CF(G= g'~\&~g'\neq g)|
	%CI= \max_{g, g' \in G~\&~g'\neq g}|CF(g)- CF(g')|
	C_\triangle = \max_{g, g' \in G~\&~g'\neq g}|F(g)- F(g')|
\end{equation}

\noindent where F refers to the \emph{concordance fraction} evaluating the group-wise correct pairwise ordering based on its respective group members and is formulated as below:

\begin{equation}
\label{equ:cf}
F(G=g) = \frac{1}{|M(G=g)|}\sum_{i|G(x_i)=g}\sum_{j\neq i} \mathbb{1}[r(x_i)> r(x_j)]
\end{equation}

%\jeremy{replace ``PP'' and other multiple character function definitions with single letters, and define those functions in the surrounding text.}

\noindent where $r(x_i)$ is the risk score assigned to the individual $x_i$ by the model, and $M$ stands for the \emph{permissible pair} whose shorter survival time is observed (the \emph{impermissible pair} is denoted as $N$ to be further discussed in Section~\ref{subsec:interplay}). The $M$ is mathematically defined as: 

\begin{equation}
\label{equ:pp}
M(G=g)= \{G(x_i)= g, x_j \neq x_i | \delta_{min(t_i, t_j)}=1 \}
\end{equation}

Specifically, CI first considers individual level pairwise comparison between model prediction and actual outcomes, \emph{i.e.}, Equation~(\ref{equ:cf}) and (\ref{equ:pp}), then measures, at the group level, \emph{i.e.}, Equation~(\ref{equ:ci}) and (\ref{equ:cf}), whether the discriminative ability of the model is fairly distributed across groups. The lower the concordance imparity score the fairer the model. Note that different from the previous definitions~\cite{keya2021equitable}, the survival time and censorship information are explicitly involved in CI to avoid important information loss and introducing substantial bias. 

\textbf{Fair calibration}. The concordance imparity depends on the risk score to measure the distinct discriminative capabilities on different demographic groups of the model. Here, we further introduce \emph{fair calibration (FC)} to evaluate probability values based prediction disparate. In contrast to the risk score based prediction disparate, which is meaningful within the context of other patients' risk scores, probability values are labels for individual patients with semantic content, thus providing stakeholders and practitioners with an additional navigation for fair decisions.   

In practice, FC starts with sorting the predicted probabilities at a specific time $t$ condition on surviving to $t$, \emph{i.e.}, $S(t|x)$, for each individual demographic group as defined by $G$. Within each demographic group, the sorted predicted probabilities are further group into deciles, \emph{i.e.}, $D = 10$ number of bins. Suppose there are 100 individuals for one certain demographic group, \emph{i.e.}, $G= g$; each bin would contain 10 individuals belong to this demographic group with their predicted probabilities in ascending order, so are other demographic groups. The final outcome of FC, denoted $C_\parallel$, is then evaluated as:

\begin{equation}
C_\parallel = 
\begin{cases}
\text{fair~calibrated,} & \text{$p(L_g(S(t|x)))\geq$ 0.05, $\forall$ g$\in$ G}\\
\text{not fair~calibrated,} & \text{otherwise}
\end{cases}
\end{equation} 

\noindent where $p(L_g(S(t|x))$ is the p-value of the Hosmer-Lemeshow goodness-of-fit test~\cite{hosmer1980goodness} for demographic group $g$ at a specific time $t$: 

\begin{equation}
L_g(S(t|x))= \sum_{i=1}^{D} \frac{({O_i}_g- {n_i}_g\bar{p_i}_g)^2}{{n_i}_g\bar{p_i}_g(1-\bar{p_i}_g)}
\end{equation} 

%n_i(1-KM_i(t))
\noindent where ${O_i}_g$, ${n_i}_g$ and $\bar{p_i}_g$ are the number of observed events, the number of individuals and the average predicted probability of bin $i$ pertaining to demographic group $g$ at time $t$, respectively. However, the exact number of observed events, ${O_i}_g$, could be unobservable when bin $i$ contains individuals censored before time $t$. To this end, the D'Agostino-Nam translation~\cite{d2003evaluation} is employed to incorporate censoring which uses the Kaplan-Meier~\cite{ranstam2017kaplan} curve estimate of bin $i$ pertaining to $g$, denoted $K_{ig}$, in place of ${O_i}_g$:   
%$\hat S_{ig}$

\begin{equation}
{O_i}_g= {n_i}_g(1-{K{_i}}_g(t))
\end{equation}

The $L_g(S(t|x))$ follows a $\chi^2_{B-1}$ distribution, and the model is fair calibrated on the condition that all p-value pertaining to each subgroup pass the test, \emph{i.e.}, a p-value greater than 0.05; otherwise, the model's predicted probability is deemed to be biased against those being underperformed, \emph{i.e.}, the model's predicted probabilities only agree with favored communities' observed event rates or frequencies of the outcome. Note that the same as concordance imparity, fair calibration explicitly involves survival time and censoring information so as to quantify unfairness in the presence of uncertainty effectively.

\subsection{Universal Survival Difference}

Random forests construct an array of base learners to increase predictive ability. It has also been extended to handle censored data, and enjoys the merits of automatically addressing the difficulties of restrictive parametric assumptions of other approaches and is capable of modeling nonlinear interactions~\cite{wang2019machine}. However, existing survival random forests aim to optimize for predictive performance and does not take fairness, which we desire to add, into consideration~\cite{ishwaran2008random}. To alleviate this limitation, we propose \emph{Survival Universal Random Forests (SURF)} to jointly consider fair data encoding and discrimination reduction for fair forests induction by: i) introducing a new accuracy-driven and fairness-oriented splitting criterion to select the potential fair splitting candidates in the presence of censorship, ii) illustrating the way to enable the construction of the new fair random forests under censorship (c.f. Section~\ref{sec:surf}).  

In the absence of censorship, the information gain and Gini impurity are normally used to measure the certainty on class labels during the induction of the tree for classification performance~\cite{han2012data}. When censorship is present, which brings uncertainty on the class labels, the survival difference between splitting nodes can be instead used to measure the impurity. To explicitly involve survival time and censoring information, we use the \emph{logrank test}~\cite{bland2004logrank} to distinguish the survival difference, denoted $S_\triangle$, between different groups:

\begin{equation}
S_\triangle = \frac{\sum_{j=1}^{k} (O_j-E_j)}{\sqrt{\sum_{j=1}^{k} V_j}} \sim N(0,1),
\end{equation}

\noindent where $O_j$, $E_j$ and $V_j$ are the observed number of events, the expected number of events and variance of the observed number of events, respectively. Splitting on candidates with a larger logrank test leads to similar survival intra child nodes and dissimilar survival inter child nodes.

We then propose a new fair splitting criterion that takes both predictive performance and fairness into consideration. We define the conjunctive criterion \emph{Universal Survival Difference (USD)}, denoted $S_\vee$, as: 

\begin{equation}
		%USD= |\log SD(X_v, X_{v'})|-\sum_{v \in dom(f)}|\log SD(X_v^g, X_v^{g'})|
		S_\vee = |\log S_\triangle(X_v, X_{v'})|-\sum_{v \in dom(f)}|\log S_\triangle(X_v^g, X_v^{g'})|
\end{equation}

\noindent where $X_v$ (or $X_{v'}$), $v$ (or $v'$) $ \in dom(f)$ are the partitions induced by feature $f$ with $X_v^g$ and $X_v^{g'}$ represent the sub-partitions further distinguished by the sensitive attribute $G$ within the partition $X_v$. In practice, the minuend focuses on the overall survival difference between different resulting tree nodes if induced by feature $f$, while the subtrahend pays attention to the survival difference between different demographic groups within each resulting tree node. 

Intuitively, USD jointly considers the predictive performance by maximizing the internode-wise overall survival difference, \emph{i.e.}, a large value of the minuend, as well as unfairness reduction by minimizing the intranode-wise survival difference among different demographic subgroups, \emph{i.e.}, the smaller subtrahend value, thus motivating fair split by giving priority to splitting candidates that ensures predictive performance while preventing from systematically underperforming on certain population simultaneously. The use of $log$ form in USD is for smoothing. In addition, such a combination of multiple factors is also advantageous when factors are not in the same scale which means one can be dominated by the other. 

This fair splitting criterion explicitly considers survival time and censorship information now gives us a means to induce fair decision trees in the presence of censorship, and thus build Random Forests under uncertainty. We emphasize that this approach to promote fair splitting has no tunable parameters as given. This is to retain the desirable property of decision trees in that they often ``just work''~\cite{han2012data}.

\subsection{Survival Universal Random Forests}
\label{sec:surf}

With the tailored fair splitting criterion specifically accounts for censoring, we now introduce a corresponding learning algorithm, \emph{Survival Universal Random Forests (SURF)}, following the general idea of random forests to generate tailored forecasts while providing fair risk predictions. The pseudocode of SURF is illustrated in Algorithm~\ref{alg:surf}.

\begin{algorithm}[!htb]
	\caption{SURF Algorithm}
	\label{alg:surf}
	\renewcommand{\algorithmicrequire}{\textbf{Input:}}
	\renewcommand{\algorithmicensure}{\textbf{Output:}}

	\newcommand{\continue}{\textbf{continue}}
	
	%\begin{spacing}{0.8}
		\begin{algorithmic}[1]
			\REQUIRE Censored and discriminated dataset $X$,\\
							%~~~~Sensitive attribute $G$,\\
							~~~~The minimum unique events for splitting $d_0$,\\
							~~~~Ensemble size B
			%		\hspace*{\algorithmicindent} Risk scores $r$,\\
			%		\hspace*{\algorithmicindent} Sensitive attribute $G$.\\ %and sensitive value $g$
			
			\ENSURE Survival universal ensemble SURF \\

			\FOR {$i \in$ $1, \cdots, B$}
				\STATE $X^{(i)}\leftarrow$ A bootstrap sample from $X$
				\WHILE {Leaf $l$ with more than $d_0$ unique events}
					\STATE $F^{(i)}\leftarrow$ A subset of the original features
					\STATE $f\leftarrow$ The highest $S_\vee(f)$ for $f \in F^{(i)}$ \\~~~~~~~based on $X^{(i)}$
					%\STATE Calculate $\overline{USD(f)}$ merit for each attribute $f \in F^{(i)}$
					\STATE Split $l$ on $f$
				 \ENDWHILE
			     \STATE SURF= SURF $\cup~ SURF_i$
		   \ENDFOR
			
			\RETURN The learned SURF
		\end{algorithmic}
	%\end{spacing}
\end{algorithm}

%\jeremy{what is the overline mean? Remove?}

The algorithm specifically works as follows: for each ensemble member of SURF, a bootstrap
sample from the censored and discriminated dataset $X$ is first selected where $X^{(i)}$ denotes the $i$th bootstrap (line 1-2). When growing the ensemble member, the candidate splitting attributes at each node are restricted to some randomly selected subset of the whole feature space to decrease the correlation between trees in the ensemble (line 4). From fairness point of view, the randomness introduced by such restriction also limits the biased correlation inherited in the data. Line 5 then performs the splitting test related calculation based on the newly proposed USD to maximize the survival difference of all demographic groups rather than certain population. When the best splitting attribute in the subset has been selected, line 6 splits the node, causing each ensemble learner to grow. As selecting which feature to split is the most computationally expensive aspect of decision tree construction, narrowing the set of features further decreases the computational requirements of SURF. SURF is in full size when all terminal nodes of each base learner, \emph{e.g}, $SURF_i$, contain less than $d_0>0$ unique event of interest (line 3).

With SURF being inducted, it is ready for fair risk predictions. SURF predicts the risk as the cumulative hazard function $H(t|x)$ to have a direct interpretation of the expected number of events, and it is the intermediate function between hazard and survival functions for direct derivation when needed. Formally, the risk score is estimated by the Nelson-Aalen estimator~\cite{borgan2014n} as:

\begin{equation}
r(t|x)= \sum_{j\leq t} \frac{d_j}{n_j}
\end{equation} 

\noindent where $d_j$ and $n_j$ represent the number of individuals experiencing the events and have not experienced the event at time $j$ respectively, and $t$ is evaluated as the last event time. In response to cases within the same node sharing identical class label in non-censoring trees, all individuals within the node of SURF have the same risk score which is used for final risk predictions.

\subsection{Interplay between Fairness and Censorship}
\label{subsec:interplay}

Armed with the previously introduced fairness notions and debiasing algorithm explicitly considering survival time and censoring information, quantifying and guaranteeing unfairness in the presence of censorship becomes feasible. In addition, it is also of practitioners and policy makers' interests to have an insight into the fairness-performance. However, due to censorship, existing fairness theoretical analyses are not applicable to understand the interplay between fairness and censorship. Here, we further provide a versatile geometric formalism to study fairness under uncertainty.

To this end, we formulate the fairness diagnostic as a fairness with uncertainty confusion tensor encoding the information needed to study the fair discriminative ability, \emph{i.e.}, concordance imparity, of the model, which provides a general perspective for understanding unfairness under uncertainty. In specific, the fairness with uncertainty confusion tensor is the stack of confusion matrices for the sensitive attribute $G$, as shown in Table~\ref{tab:censoredFairness-confusion}.  For simplicity, we assume that $G$ is a binary attribute, \emph{i.e.}, $G\in\{0, 1\}$.

\begin{table}[!htbp]
	\caption{The fairness with uncertainty confusion tensor, showing the two planes corresponding to the confusion matrix for each of the favored (G = 1) and deprived groups (G = 0).} 
	\centering
	\begin{tabular}{ccc}
		\toprule
		G = 0  &  r $>$ r'  & r $<$ r'  \\
		\midrule
		t $>$ t'  \& $\delta$' = 0  & ${N}_0^a$    & ${N}_0^b$  \\
		t $>$ t'  \& $\delta$' = 1   & ${M}_0^a$   & ${M}_0^b$   \\
		t' $>$ t  \& $\delta$ = 0  & ${N}_0^c$    & ${N}_0^d$  \\
		t' $>$ t  \& $\delta$ = 1  &  ${M}_0^c$    & ${M}_0^d$   \\
		\bottomrule
		\vspace{+0.001cm}
	\end{tabular}
	
	\begin{tabular}{ccc}
		\toprule
		G = 1  	 &  r $>$ r'  & r $<$ r'  \\
		\midrule
		t $>$ t'  \& $\delta$' = 0  & ${N}_1^a$    & ${N}_1^b$  \\
		t $>$ t'  \& $\delta$' = 1   & ${M}_1^a$   & ${M}_1^b$   \\
		t' $>$ t  \& $\delta$ = 0  & ${N}_1^c$    & ${N}_1^d$  \\
		t' $>$ t  \& $\delta$ = 1  &  ${M}_1^c$    & ${M}_1^d$   \\
		\bottomrule
	\end{tabular}
	\label{tab:censoredFairness-confusion}
\end{table}

Let us denote the elements of fairness with uncertainty confusion tensor as ${N}_G^a$, ${N}_G^b$, ${N}_G^c$ and ${N}_G^d$ as well as ${M}_G^a$, ${M}_G^b$, ${M}_G^c$ and ${M}_G^d$ (abbreviations in align with Equation~\ref{equ:pp}), each element with subscripts indicating the value of $G$. We further denote $P_G= {M}_G^a+ {M}_G^b+ {M}_G^c+{M}_G^d$, $I_G= {N}_G^a+ {N}_G^b+ {N}_G^c+{N}_G^d$ and $C_G= {M}_G^b+ {M}_G^c$ be the number of pairwise individuals in each group $G\in\{0, 1\}$ pertaining permissible, impermissible and concordant pairs, respectively. Assume $P_G$, $I_G$ and $C_G$ are known constants. Unraveling the fairness with uncertainty confusion tensor, we can formulate concordance imparity as:
 
\begin{equation}
	%CI= \frac{{PP}_0^b+{PP}_0^c}{P_0} - \frac{{PP}_1^a+{PP}_1^d}{P_1}
	C_\triangle = \left|\frac{C_0}{P_0} - \frac{C_1}{P_1}\right|
\end{equation}

We show below that how fair/unfair the model could appear in the presence of censorship, \emph{i.e.}, how censorship could affect fairness notion calculation, based on ``floor'' outcome and ``ceiling'' outcome possible. 

Let us first consider the \emph{``floor'' outcome}--  all impermissible pairs, \emph{e.g.}, $I_G$, become permissible with all censored individuals experience immediate event of interest, \emph{i.e.}, $\delta$ becomes 1 for those individuals with $\delta=0$ and their censored time $t$ now becomes the actual event time. Under this floor circumstance the CI becomes:

\begin{equation}
\lfloor C_\triangle \rfloor = \left|\frac{C_0+ I_0^{con}}{P_0+I_0} - \frac{C_1+ I_1^{con}}{P_1+I_1}\right|
\end{equation}

%Note that all previously impermissible pairs, \emph{e.g.}, $I_G$, also become permissible. 
\noindent where $I_0^{con}= N_0^b+N_0^c$ and $I_1^{con}= N_1^b+N_1^c$ representing the additional concordant pairs under this floor circumstance. Now we are ready to discuss different sub-scenarios possible affecting fairness notion calculation to provide practitioners and stakeholders more insights. 

\begin{itemize}
	\item \emph{Sub-scenario 1: $\frac{C_0}{P_0}= \frac{I_0^{con}}{I_0}$ and $\frac{C_1}{P_1}= \frac{I_1^{con}}{I_1}$.} 
	
	In this case, $\lfloor C_\triangle \rfloor= \left|\frac{C_0+\frac{C_0*I_0}{P_0}}{P_0+I_0}- \frac{C_1+\frac{C_1*I_1}{P_1}}{P_1+I_1}\right|= \left|\frac{C_0(1+\frac{I_0}{P_0})}{P_0+I_0}- \frac{C_1(1+\frac{I_1}{P_1})}{P_1+I_1}\right|= \left|\frac{C_0(1+\frac{I_0}{P_0})}{P_0(1+\frac{I_0}{P_0})}- \frac{C_1(1+\frac{I_1}{P_1})}{P_1(1+\frac{I_1}{P_1})}\right|=\\ ~\left|\frac{C_0}{P_0}- \frac{C_1}{P_1}\right|$
	
	That is to say CI stays unchanged in this case. In practice, this suggests that the independence of model prediction with censorship. Practitioners and stakeholders can therefore interpret the level of unfairness of the model under consideration with more confidence. 
	\item \emph{Sub-scenario 2: $\frac{C_0}{P_0}= 0$ and $\frac{C_1}{P_1}= 1$.} 
	
	In this case, the uncertainty affect the fairness calculation the most and could lead to the most deviated CI score from the original CI score in the presence of censorship. Sufficient care must be taken to ensure the extra unfairness possible as the result of the censorship when deploying social sensitive applications.  
	
	\item \emph{Sub-scenario 3: $\frac{C_0}{P_0}= 0$ and $\frac{C_1}{P_1}= 0$.}  
	
	In contrast to the sub-scenario 2, the uncertainty, in this case, does affect the fairness calculation but uniform in direction. Practitioners and stakeholders can have this in mind considering real-world's socially sensitive impacts. 
	
	\item \emph{Sub-scenario 4: assume certain distribution with $\frac{C_0}{P_0}$ and $\frac{C_1}{P_1}$, for example the binomial distribution.} 
	
	The assumed distribution on censorship in conjunction with existing CI score in the presence of censorship can therefore provide additional confidence interval with respect to the range of the CI score in the presence of censorship. 
\end{itemize}

In contrast to the previous ``floor'' outcome, the second \emph{``ceiling'' outcome} also presents all impermissible pairs become permissible but all censored individuals now have an infinite event time of interest instead, \emph{i.e.}, $\delta$ also becomes 1 for those individuals with $\delta$ = 0 but their censored time $t$ becomes infinite instead. Under this floor circumstance the concordant elements of $\lceil C_\triangle \rceil$ change to the other way around, \emph{i.e.}, $N_0^{con}= N_0^a+N_0^d$ and $N_1^{con}= N_1^a+N_1^d$ with all previously impermissible pairs, \emph{e.g.}, $I_G$, become permissible as well. In addition, the censorship of longer survival time also makes difference and should be considered to form the new uncertainty tensor. Other than these, the previous discussion on ``floor'' outcome still applies and can be analyzed similarly to help the practical application deployment. 

%\jeremy{The table hangs over the right margin}

\begin{table}[!htbp]
	\footnotesize
	%\normalsize
	%\small
	\caption{Summary of datasets used in experiments.} 
	\begin{tabular}{cccccc}
		\toprule
		\diagbox{Characteristics}{Datasets}  & ROSSI & COMPAS  & KKBOX  \\
		\midrule
		%\multirow{5}{*}{Rossi} 
		Sample \# 				& 432     &  10,325   & 2,814,735   \\
		Censored \# 		  & 318  	 & 7,558     & 975,834   \\
		Censored Rate		& 0.736   & 0.732     & 0.347   \\
		Feature \# 				& 9           & 14          & 18 \\
		Sensitive Attribute	& race      & race       & gender \\
		Sensitive Value		 &black     &   \begin{tabular}[c]{@{}c@{}}African\\ American \end{tabular}    & female \\ \bottomrule
	\end{tabular} 
	\label{tab:dataset_info}
\end{table}

\begin{table*}[!htbp]
	%\footnotesize
	%\normalsize
	\small
	\centering
	\caption{Evaluation results of different models with the best results marked in bold.} 
	\begin{tabular}{ccccccc}
		\toprule
		Datasets & \diagbox{Method}{Metrics} & CI\% & Fair Calibration & C-index\% & Brier Score\% & Time-dependent AUC\%\\
		\midrule
		\multirow{6}{*}{ROSSI} 
		& IDCPH  &  15.31 & Not fair calibrated  		  & 52.28  & 18.73 &  77.32 \\
		& GDCPH &  9.32  & \textbf{Fair calibrated}    & 59.34 & 22.87 & 78.51 \\
		& CPH     & 11.43  & Not fair calibrated 			& 64.24 & 17.67 & 77.12\\
		& RSF      & 16.53 & Not fair calibrated 			& 65.56 & 15.12 & 79.32\\
		& DeepSurv  & 12.32 & Not fair calibrated 		& 66.67  & 14.71 & 77.17\\
		& SURF   & \textbf{5.8}  &  \textbf{Fair calibrated}   & \textbf{69.33} & \textbf{12.33}& \textbf{79.39}\\
		\midrule
		\multirow{6}{*}{COMPAS}  
		& IDCPH      & 25.18 & 	Not fair calibrated         & 62.16  & 25.03 & 63.78 \\
		& GDCPH   	& 11.77  &	\textbf{Fair calibrated}  & 72.16 & 16.32 &  66.21  \\
		& CPH 			& 22.43 &  Not fair calibrated	     & 69.24 & 20.35 &  65.15 \\
		& RSF 		 	& 25.32 &   Not fair calibrated       & 72.61   & 15.62 &  71.76\\
		& DeepSurv  & 16.72 &  Not fair calibrated         & 75.12 & \textbf{13.42}& 71.83\\
		& SURF		& \textbf{8.31}  &  \textbf{Fair calibrated}& \textbf{76.43}  & 13.72 & \textbf{72.03}\\
		\midrule
		\multirow{6}{*}{KKBOX} 
		& IDCPH   &  17.79 & Not fair calibrated		   & 72.61 				& 21.23 &  69.73 \\
		& GDCPH   & 14.98 & \textbf{Fair calibrated}   & 79.45  		   & 19.92 &  73.03 \\
		& CPH      & 18.91 &  Not fair calibrated    		 &  80.02 			 & 18.17 & 72.95 \\
		& RSF      & 21.14 &  Not fair calibrated  			  &  82.32 			   & 14.24 &  78.18\\
		& DeepSurv  & 20.66  &  Not fair calibrated 	& \textbf{83.01} & 14.33  & 80.71\\
		& SURF   & \textbf{12.57} & \textbf{Fair calibrated} & 82.72 & \textbf{13.21}& \textbf{81.21} \\
		\midrule
	\end{tabular}
	\label{tab:effectiveness}
\end{table*}

\section{Experiments} 
\label{sec:expt}

%This section....

\subsection{Datasets}

We validate our proposed SURF on three real-world censored datasets with socially sensitive concerns and diverse characteristics\footnote{The code and data are publicly available at\\ \indent \url{https://github.com/vanbanTruong/censoredFairness}}. Table~\ref{tab:dataset_info} provides a summary description of them. Note that survival time and censoring information are explicitly included in our study to specifically account for censorship. 

The \textbf{ROSSI} dataset pertains to persons convicted then released from Maryland state prisons, who were followed up for one year after release~\cite{fox2012rcmdrplugin}. Among the total of 432 released convicts engaging in an experimental treatment, half of them were given financial aid and the rest of them did not receive aid. The task is then to predict the reoffending risk score of convicted criminals as described by 9 features, including the sensitive attribute ``race'' with black being the deprived group and other being the favored group.

The landmark algorithmic unfairness \textbf{COMPAS} dataset has a similar learning task but
is considerably larger with 10,325 convicts from Broward County~\cite{angwin2016there}. The COMPAS dataset consists of 14 features with ``race'' being the sensitive attribute and the  African American defining the deprived group in contrast to other as the favored group. The COMPAS dataset also shares a similar 73\% censored rate with the ROSSI dataset.

The \textbf{KKBOX} dataset is created from the WSDM-KKBOX's Churn Prediction Challenge 2017~\cite{kvamme2019time}. Specially, the task is to predict a user's risk score of canceling his/her streaming music subscription from KKBOX. Compared with the previous two datasets, this is the largest number dataset we experiment with along with a lower censored rate of 34.7\%. The total 2,814,735 users are described using the same 18 features developed by the winning Kaggle team, and gender is used as the sensitive attribute with female being the sensitive value.

\subsection{Experimental Setup}
We compare our method with the representative survival models and recent survival models with fairness constraints based on a set of typical survival analysis evaluation metrics as well as our proposed fairness notions considering censorship measuring their fair risk prediction capabilities. 

\textbf{Evaluation Metrics.}
In addition to the proposed fairness measures considering censorship (c.f. Section~\ref{sec:fairMetrics}), in our evaluation we also report the following widely used survival analysis evaluation metrics: i) the \emph{C-index}, which is an accuracy measure equals to the area under ROC curve (AUC) in the absence of censorship~\cite{harrell1982evaluating}, ii) the \emph{Brier score} in measuring the mean squared error between the probability estimations assigned to possible outcomes and the actual outcome~\cite{brier1951verification}, iii) the \emph{Time-dependent AUC}, which is a function of time testing the discriminative capability of a model when distinguishing individuals experienced the event of interest from those have not experienced till time $t$~\cite{chambless2006estimation}. Among them, while the model with a lower Brier score is desired, the higher C-index and Time-dependent AUC scores, the better.

\textbf{Comparison Methods.} The proposed SURF is designed to address highly underexplored discrimination in the presence of censorship. To evaluate its design, we consider baselines from the following four perspectives: i) two recent proportional hazards assumption based fair survival models IDCPH and GDCPH~\cite{keya2021equitable}, which are the only works for fair survival analysis problem to the best of our knowledge (note that only the most competitive ones are discussed among different variants proposed therein), ii) the mostly widely used CPH for survival analysis~\cite{cox1972regression}, iii) the state of the art random forests on survival analysis RSF~\cite{ishwaran2008random} and deep neural network based DeepSurv~\cite{katzman2018deepsurv} as additional baselines. Other competing fairness methods are not considered as none of them is capable of addressing fairness in the presence of censoring. 

%\jeremy{Remove the white spaces around the graphs so we can read the axes and the legend.}

\begin{figure*}[!htbp]
	\caption{The binned fair calibration for deprived community, graphing the predicted probabilities against the true probability in deciles. The color of the bar represents different methods with the red bar representing the true probability observed by KM.}
	\centering
	%\vspace{-2mm}
	%\makebox[0.6\textwidth][l]{
	\subfigure[ROSSI]{
		\includegraphics[height=0.25\textheight,width=0.3\textwidth]{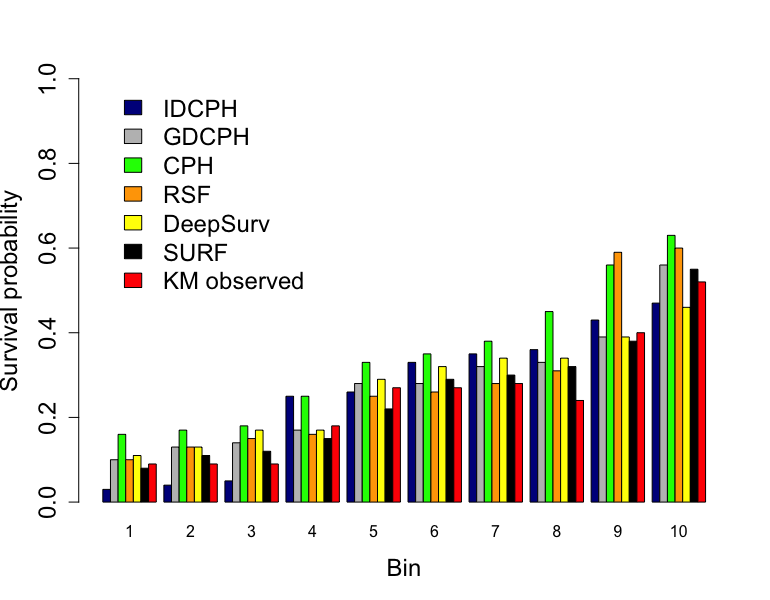}
	}
	\subfigure[COMPAS]{
		\includegraphics[height=0.25\textheight,width=0.3\textwidth]{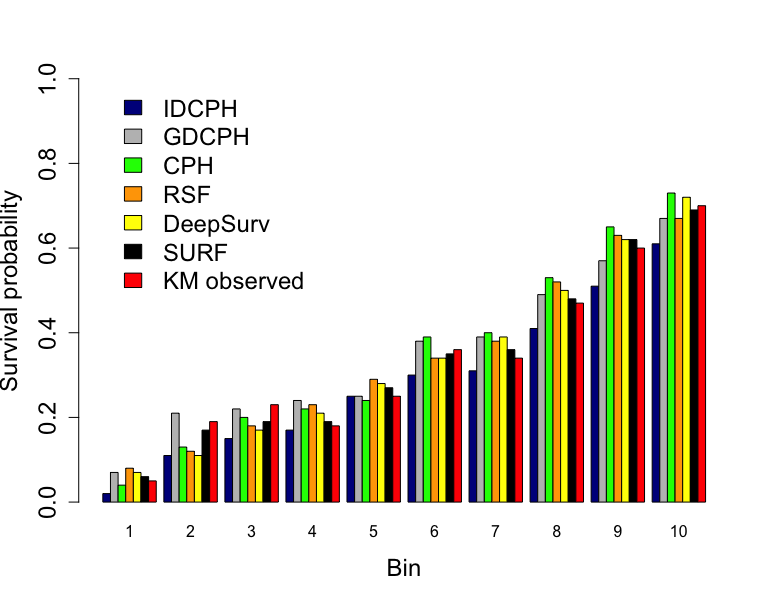}
	}%
	\subfigure[KKBOX]{
		\includegraphics[height=0.25\textheight,width=0.3\textwidth]{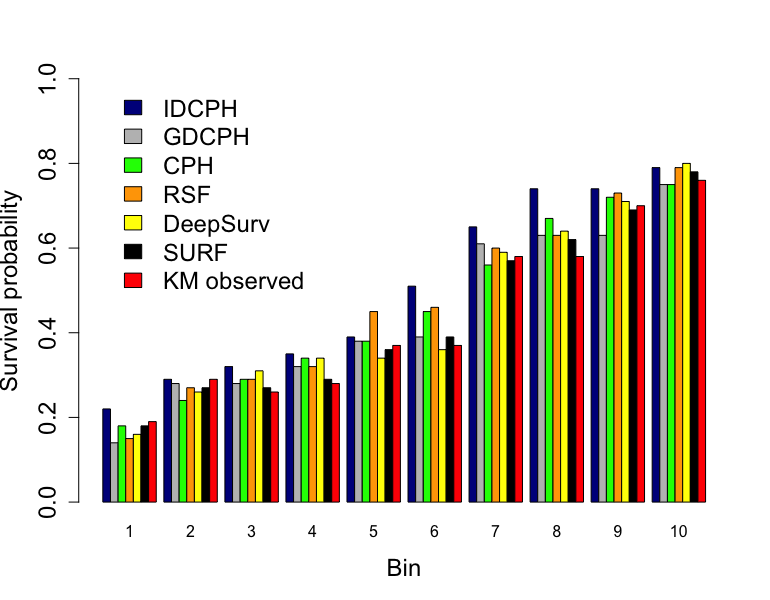}
	}%
	
	%}

	\label{fig:fairCalibration}
\end{figure*}

\subsection{Results}

We present both quantitative and qualitative experiment results that illustrate and confirm the utility of SURF in making fair decisions under uncertainty. 

\textbf{Fair risk predictions.} A set of fairness metrics in the presence of censorship and survival analysis evaluations are tested on different models with the 5-fold cross validation results summarized in Table~\ref{tab:effectiveness}.

\begin{table*}[!htbp]
	%\footnotesize
	%\normalsize  
	\small
	\caption{Predictive performance confusion matrix of SURF.} 
	\centering
	\begin{tabular}{ccccccccccccc}
		\toprule
		\multirow{2}{*}{Datasets}& \multicolumn{2}{c}{\begin{tabular}[c]{@{}c@{}}C-index\%\\ deprived\end{tabular}} & \multicolumn{2}{c}{\begin{tabular}[c]{@{}c@{}}C-index\%\\ favored\end{tabular}}& \multicolumn{2}{c}{\begin{tabular}[c]{@{}c@{}}Brier Score\%\\ deprived\end{tabular}} & \multicolumn{2}{c}{\begin{tabular}[c]{@{}c@{}}Brier Score\%\\ favored\end{tabular}}& \multicolumn{2}{c}{\begin{tabular}[c]{@{}c@{}}Time-dependent\\ AUC\% deprived\end{tabular}} & \multicolumn{2}{c}{\begin{tabular}[c]{@{}c@{}}Time-dependent\\ AUC\% favored\end{tabular}}\\ 
		\cline{2-13}
		&  SURF-  & SURF &SURF- & SURF &  SURF- & SURF &  SURF- & SURF &  SURF- & SURF &  SURF- & SURF\\
		\midrule
		ROSSI   	 & 51.71    & 65.32    & 74.24  & 71.12  & 21.03  &  16.23  & 9.87 & 10.17  & 69.98 & 73.65 & 84.87 & 82.17 \\
		COMPAS  & 54.82    &  70.22   &  80.14  & 78.53 &  18.76  & 16.12 &  7.66 & 11.65 & 62.81 & 65.17  & 77.62 & 75.24\\
		KKBOX     &  64.53    & 72.89   &  85.67  & 85.46 &  18.87  & 15.17 & 7.16  & 9.12 & 72.31 & 77.45 & 85.87 & 84.66 \\
		\bottomrule
	\end{tabular}
	\label{tab:confusionmatrix}
\end{table*}

As can be seen clearly our new SURF dominates all other baselines in terms of minimizing discrimination while maintaining a competitive predictive performance, which verifies the necessity of its debiasing design while accounting for censorship. On the other hand, the lack of consideration for survival time and censoring information as well as task-specific similarity metric among individuals result in the inferior performances of other baselines, although they are proposed for fairness in censoring settings. This also verifies our previous discussion that fairness in the presence of censorship cannot be trivially solved by a simple combination of existing techniques.

As p-values are not intended to be ranked, the fair calibration cannot rank models besides suggesting whether one model is calibrated across different demographic communities. We therefore further visualize the predicted probabilities against the true probability observed by KM to qualitatively evaluate the fairness of such predictions, as shown in Figure~\ref{fig:fairCalibration}. Note that for space constraint, only the fair calibration plots for deprived community is presented. As we can see, the dark bars, are roughly the
same height as the red ones, suggesting SURF's predicted probabilities are representative of the deprived community's true probabilities.

In addition, we also perform an ablation study, testing whether the debiasing component of SURF worked to benefit the improved performance of the deprived group. To this end, we further break down the predictive performance by the sensitive value that defines the deprived community and favored community and with and without our fairness constraints (\emph{e.g.}, SURF and SURF-). The predictive performance confusion matrix are shown in Table~\ref{tab:confusionmatrix}. As one can see, SURF attends to the deprived community with improved predictive ability and characterization, which verifies its anti-discrimination capability. In addition, the improved overall predictive performance also shows the merit of such anti-discrimination design for fair risk predictions as a whole.

\begin{figure}[!htbp]
	\centering
	\caption{The time complexity comparison of all the methods.}
	\includegraphics[height=0.25\textheight,width=0.4\textwidth]{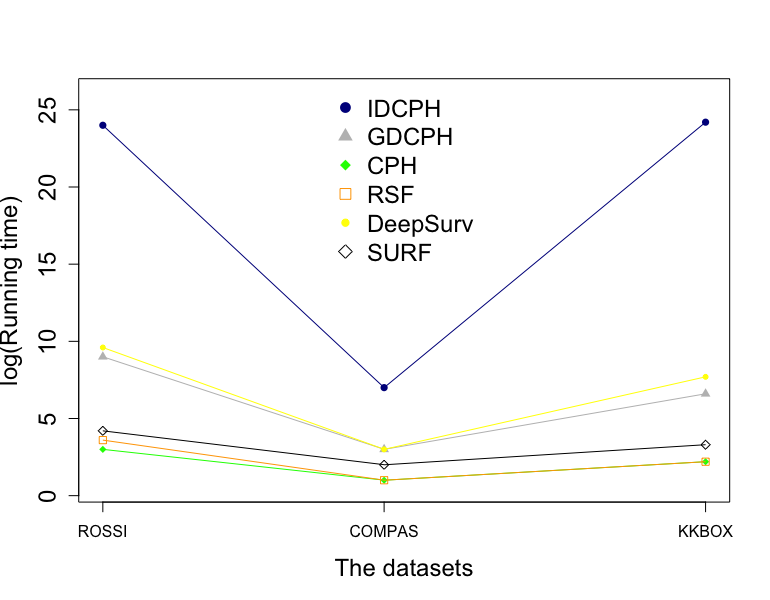}
	\label{fig:scalability}
\end{figure}

\textbf{Time Complexity.} We illustrate and analyze the time complexity of our proposed SURF against other baseline methods, as shown in Figure~\ref{fig:scalability}, where the y-axis is the logarithm of the runtime in seconds for ROSSI dataset and minutes for the other two datasets. As one can see, SURF enjoys the merit of computational efficiency as it follows the idea of tree based method to decrease the computational requirements in practice, while the other baseline IDCPH becomes computational expensive, thus limiting its applicability.

\section{Conclusion}
\label{sec:con}

This work is motivated by the increasing attention on the issue of discriminatory AI behaviors, with the aim to provide fair and accuracy predictions in the presence of censorship. Different from existing fair survival methods, we propose two censored-specific fairness notions and a new debiasing algorithm to specifically account for fairness under censorship. The positive results of conducted experiments show the anti-discrimination capability of our proposed method. The proposed technique is expected to be versatile in alleviating bias in various socially sensitive applications ({\emph{e.g.}, the allocations of health resources, personalized marketing and recidivism prediction instrument).

\bibliographystyle{IEEEtran}
\bibliography{typeinst}

\end{document}